\pdfoutput=1

\documentclass[11pt]{article}

\usepackage{emnlp2021}

\usepackage{times}
\usepackage{latexsym}

\usepackage[T1]{fontenc}

\usepackage{subfig}
\usepackage{graphicx}

\usepackage{tikz} \usetikzlibrary{calc} \tikzset{>=latex}

\usepackage{microtype}

\title{ASQ: Automatically Generating Question-Answer Pairs using AMRs}

\author{Geetanjali Rakshit and Jeffrey Flanigan \\
Computer Science and Engineering Department\\
UC Santa Cruz \\
\texttt{\{grakshit,jmflanig\}@ucsc.edu} }
 
\begin{document}
\maketitle
\begin{abstract}
We introduce ASQ, a tool to automatically mine questions and answers from a sentence using the Abstract Meaning Representation (AMR). Previous work has used question-answer pairs to specify the predicate-argument structure of a sentence using natural language, which does not require linguistic expertise or training, and created datasets such as QA-SRL and QAMR, for which the question-answer pair annotations were crowdsourced. Our goal is to build a tool (ASQ) that maps from the traditional meaning representation AMR to a question-answer meaning representation (QMR). This enables construction of QMR datasets automatically in various domains using existing high-quality AMR parsers, and provides an automatic mapping AMR to QMR for ease of understanding by non-experts. A qualitative evaluation of the output generated by ASQ from the AMR 2.0 data shows that the question-answer pairs are natural and valid, and demonstrate good coverage of the content. We run ASQ on the sentences from the QAMR dataset, to observe that the semantic roles in QAMR are also captured by ASQ.
We intend to make this tool and the results publicly available for others to use and build upon.

\end{abstract}

\section{Introduction}
\label{introduction}

In the English language, predicate-argument structure of a sentence plays a key role in understanding its meaning. The idea of annotating predicate-argument structure been explored in the past in the creation of linguistic resources such as PropBank~\cite{palmer2005proposition}, FrameNet~\cite{baker1998berkeley}, OntoNotes~\cite{pradhan2007ontonotes}, NomBank~\cite{meyers2004nombank}, Groningen Meaning Bank (GMB)~\cite{bos2017groningen}, inter alia. Relational semantics at the sentence level has been encoded in the Abstract Meaning Representation (AMR) ~\cite{banarescu2013abstract} using PENMAN notation. In contrast, \citet{he-etal-2015-question} introduced the idea of using question-answer pairs to represent the semantic roles in a sentence, where each predicate-argument relationship is expressed as a question-answer pair. This idea was further explored in large scale QA-SRL parsing~\cite{fitzgerald2018large}, and QAMR~\cite{michael2017crowdsourcing}.

Question-answer meaning representations (QMRs) are appealing because they are intuitive and understandable to non-experts. They have been shown to closely match the structure of traditional predicate-argument structure annotation schemes (PropBank, FrameNet, NomBank), without the need for predefined frames or thematic role ontologies. However, QMRs contain the complexities and ambiguities of natural language. On the other hand, standard MRs are unambiguous, and are widely used in downstream tasks, but are not understandable to non-experts.

\begin{figure}
\centering
\includegraphics[width=0.5\textwidth]{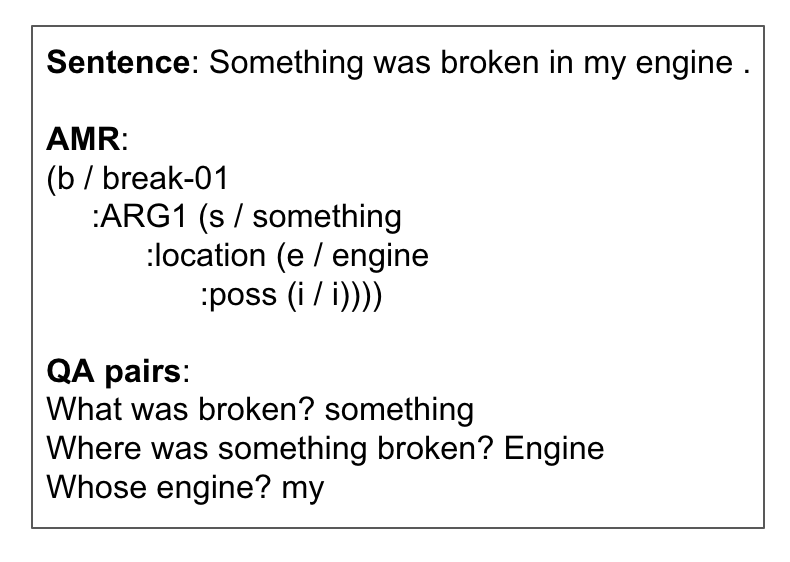}
\caption{Example from ASQ}
\label{amr-qa-example}
\end{figure}

We would like to be able to map automatically between standard MRs and QMRs, to be able to present the same information either in a format understandable to computers, or in a format understandable to non-expert humans. We take a first step in this direction and demonstrate this is possible by creating an automatic system to map AMR to a QMR.
We introduce AMR Sourced Questions (ASQ), a tool to automatically mine questions and their answers from the AMR for a sentence. Figure~\ref{amr-qa-example} shows an example of representing an AMR as question-answer (QA) pairs. We handcraft templates to convert the AMR into compact, non-redundant QA pairs, with good coverage.

We run ASQ on the sentences from the QAMR dataset, to show that the semantic roles in QAMR are also captured by ASQ.

We show that outputs from ASQ have good coverage with minimal redundancy, while being fast and requiring minimal supervision, to produce good quality QA pairs. The strength of ASQ comes from the quality of templates and that of the AMRs.

The main contributions of this work are:
\begin{itemize}
\item ASQ: a tool to automatically obtain semantic information in the form of question-answer (QA) pairs, given a sentence and its AMR. To the best of our knowledge, generating question-answer pairs using AMR has not been done before.
\item A collection of rich templates for question generation, based on AMR relations.
\item A question answering dataset built from AMR 2.0, containing the range of semantics expressed by AMRs.
\item Additional question answering annotations, using ASQ and an automatic AMR parser, on sentences from the QAMR dataset.

\end{itemize}

\section{Related Work}

\citet{he-etal-2015-question} introduced the QA-SRL task to specify the predicate argument structure of a sentence using natural language in the form of question-answer pairs. The QA-SRL dataset was expanded to a large scale in~\citet{fitzgerald2018large}. \citet{michael2017crowdsourcing} introduced QAMR, which differs from QA-SRL in focusing on all words in the sentence (and not just verbs). These resources are shown to capture the same kinds of predicate-argument structures as PropBank~\cite{palmer2005proposition}, FrameNet~\cite{baker1998berkeley}, etc. Meaning representations prior to these include UDS~\cite{white2016universal}, UCCA~\cite{abend2013universal}, GMB~\cite{basile2012developing}, etc.

While both QAMR and QA-SRL are manually created, QA-SRL defines a limited grammar and templates to help annotators frame questions, and fast track the annotation process. Our work is also highly template-based, but we use richer and more number of templates, and we do not define a grammar for it. QAMR, on the other hand, doesn't use templates and allows free-form questions instead.

\section{Background}

\subsection{Abstract Meaning Representation}
Abstract Meaning Representation is a semantic representation of a whole sentence to convey its logical meaning ~\cite{banarescu2013abstract}. It draws heavily from PropBank~\cite{palmer2005proposition}, while adding more semantic information of its own. AMRs do not specify information like tense and aspect.

AMRs are represented as rooted, labeled graphs, which may be written in the graph format, or as a string in the PENMAN notation, or, as logical triples. Figure~\ref{amr-qa-example} shows a sentence and its AMR.
Each relation in the AMR graph has a \textbf{relation} name, a \textbf{concept}, and a \textbf{variable}. For example, in figure~\ref{amr-qa-example}, the variables are: \textit{b, s, e, i}; the corresponding \textit{concepts} are: \textit{break-01}, \textit{something}, \textit{engine}, \textit{i}; and the \textit{relations}: \textit{ARG1}, \textit{location} and \textit{pos}. The root doesn't have a relation name. The numeral \textit{-01} in \textit{break-01} denotes the sense of the verb \textit{break} from PropBank.

Sentences with the same meaning are expected to have the same AMR (although there isn’t always a single unique AMR for a sentence). One AMR can therefore have multiple surface realizations, all meaning the same.
The relations express how words in the sentence are related. These include \textbf{core roles}, i.e., frame arguments (following PropBank) such as \textit{arg0, :arg1, :arg2, :arg3, :arg4, :arg5}; and \textbf{non-core roles} such as general semantic relations like \textit{:accompanier, :age, :beneficiary, :cause;} relations for quantities like \textit{:quant, :unit, :scale;} relations for date-entities \textit{:day, :month, :year, :weekday, :time;} and relations for lists \textit{:op1, :op2, :op3, :op4}.
There are about 100 relations listed in the AMR specifications. These relations are the key to generating question-answer pairs in our approach, as described in the subsequent sections.

\begin{figure*}
\centering
\includegraphics[width=0.9\textwidth]{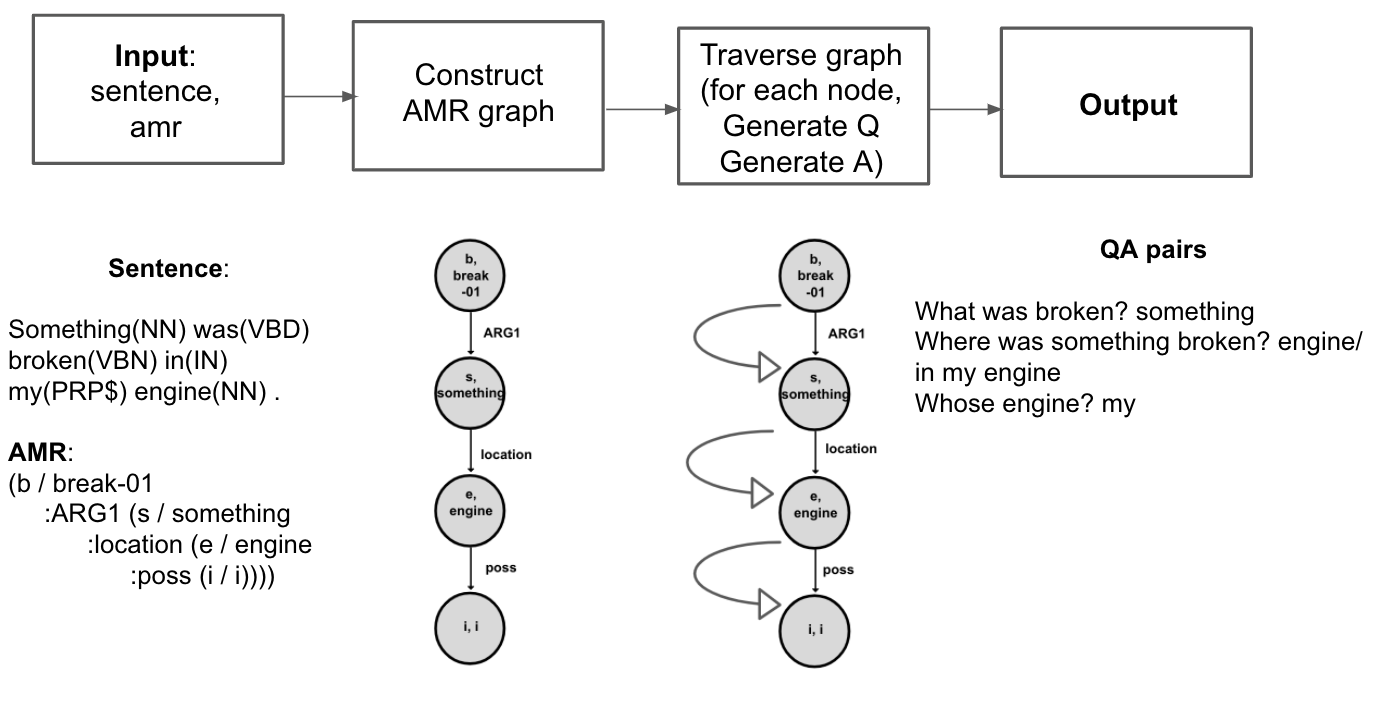}
\caption{System Overview}
\label{fig-amr-qa-pipeline}
\end{figure*} 

\section{Approach}
\label{sec:approach}
Our approach uses the sentence text and its AMR to produce question-answer pairs. 
We create a resource of handcrafted templates to generate questions from, described in detail in \S\ref{sec:template-creation}. We have at least one template for each relation. These templates are transformed into a suitable question to be answered from the sentence under consideration.

The overall system pipeline is shown in figure ~\ref{fig-amr-qa-pipeline}. We read the AMR as a string in the PENMAN format, and construct an AMR graph over it (\S\ref{sec:amr-graph-construction}), where each node is labeled with a \textit{variable} name and \textit{concept}, and the edges are labeled with \textit{relations}. 

\subsection{Template Creation}
\label{sec:template-creation}
For the relations in the AMR specifications~\cite{banarescu2012abstract}, we handcraft one or more templates. These templates are then transformed to construct a suitable question that can be answered by the sentence. The distinction between the two kinds of AMR relations, core and non-core, is based on whether or not the predicate information is essential to understand the meaning of the relation of the word to its parent word, or vice versa. Therefore, the construction of templates for each kind of role is handled differently, as described next. 
Table~\ref{tab-stats-temp} shows some statistics about the templates created.

\begin{table}
\centering
\begin{tabular}{ll}
\hline
\textbf{Core roles} & \\ \hline \hline
Number of templates & 483 \\ \hline
Avg number of templates & \\
for each relation & 7.3 \\ \hline
\textbf{Non-core roles} & \\ \hline \hline 
Number of templates & 385\\ \hline
Avg number of templates & \\
 for each relation & 2.5 \\ \hline
\end{tabular}
\caption{\label{tab-stats-temp}
Statistics about Templates
}
\end{table}

The core roles in AMR come from the numbered argument roles in PropBank~\cite{palmer2005proposition}. The meaning of the arguments depend on the predicate, and could have completely different meanings for the same core relation but different predicates. 
The predicates themselves carry sense annotations from PropBank.

Designing templates for all the predicate-argument combinations is a task in itself. As a workaround, we use the mappings in Semlink~\cite{palmer2009semlink} to associate each numbered argument with a \textbf{thematic role} in VerbNet~\cite{kipper2000class}. We then create templates for each thematic role in VerbNet. When we have to choose templates based on a numbered argument and predicate, we can find the thematic role that the argument maps to, and use the templates for that role. When creating these templates, we also store the tense of the verb required to convert the template into a valid question. This helps to narrow down on the number of choices when selecting a template to construct questions.
Table~\ref{tab-core-templates} shows templates for some of the core-roles.

\paragraph{Non-core Roles}
For the non core roles, the \textit{relation} name is sufficient to characterize templates for it. We do not use predicate information to create these templates. Table~\ref{tab-non-core-templates} shows some of the templates for the core-roles.

Some templates have multiple blanks to be filled. We store the accepted parts-of-speech tag information for each blank.

\begin{table}
\setlength{\tabcolsep}{0pt} 
\centering
\begin{tabular}{ccll}
\hline
\textbf{Rel} & & \textbf{Tense} & \textbf{Template} \\ \hline
ARG0 &~~~& past, pres. & Who \_\_\_\_\_ ? \\
     & & past, pres. & What \_\_\_\_\_  ? \\  
     & & pres. & Who is \_\_\_\_\_  ?  \\ 
     & & pres. & What is \_\_\_\_\_  ?  \\
     & & pres. & Who are \_\_\_\_\_  ?  \\ 
     & & pres. & What are \_\_\_\_\_  ? \\  
     & & past & Who was \_\_\_\_\_  ?  \\ 
     & & past & What was \_\_\_\_\_  ?  \\
     & & past & Who were \_\_\_\_\_  ?  \\ 
     & & past & What were \_\_\_\_\_  ?  \\ 
     & & future & What will \_\_\_\_\_  ?  \\ 
     & & future & Who will \_\_\_\_\_  ?  \\ \hline
ARG1 & & past & Who does someone \_\_\_\_\_ ?  \\
     & & past & What does someone \_\_\_\_\_ ?  \\ 
     & & pres. & What is someone \_\_\_\_\_ ?   \\ 
     & & future & What will someone be \_\_\_\_\_ ?   \\ \hline 
ARG2 & & pres. & What is something  \_\_\_\_\_ from ?  \\
     & & pres. & What is something  \_\_\_\_\_ of ?  \\ \hline 
ARG3 & & pres. & Who is something \_\_\_\_\_ for ?  \\
     & & pres. & What is something  \_\_\_\_\_ for ?  \\ \hline 
  \hline 
\end{tabular}
\caption{\label{tab-core-templates}
Some templates for the core roles for the predicate \textit{make}. ARG0 is the agent, ARG1 is the product, ARG2 is the material, ARG3 is the beneficiary. Tense is the tense of the generated surface form of make.
}
\end{table}

\begin{table*}
\centering
\begin{tabular}{cccl}
\hline
 \textbf{Relation} & \textbf{POS tag} & \textbf{Tense} & \textbf{Template} \\ \hline
location & verb & present & Where is  \_\_\_\_\_  ?  \\
 & verb & past & Where was  \_\_\_\_\_  ?  \\
 & verb & past & Where did someone  \_\_\_\_\_  ?  \\  \hline
mod & verb & past & Which \_\_\_\_\_  ?  \\
 & verb & past  & What type of \_\_\_\_\_  ? \\  \hline
 frequency & verb, noun & present & How many times someone  \_\_\_\_\_   \_\_\_\_\_ ? \\
 & verb, noun & present & How many times something  \_\_\_\_\_   \_\_\_\_\_ ?  \\ \hline
\end{tabular}
\caption{\label{tab-non-core-templates}
Templates for Non-core Roles
}
\end{table*}

\subsection{AMR Graph Preprocessing}
\label{sec:amr-graph-construction}

We construct a tree over the AMR graph by using the tree structure defined by the PENMAN notation of the annotation.
Once this tree is constructed, we perform some steps of preprocessing to avoid redundancy and keep cohesive bits of information together by condensing the children nodes into the parent. For example, when the concept for a node is \textit{temporal-quantity}, we replace it with the concepts from children nodes with relations \textit{:quant} and \textit{:unit} to get a concept like "1 year". We do this for concepts like \textit{date-entity, distance-entity, area-entity, volume-entity}. For leaf nodes with the relation \textit{:op}, we combine the concepts from children at the same level with relation \textit{:op\{x\}} where x can be 1, 2, 3, etc, which helps to extract multi word proper nouns as a single concept. We ignore relations like \textit{:polarity, :wiki, ,:polite, :polite-of, :mode} for question generation since they often do not contain significant information to pose questions about. 
 
We do a depth-first pre-order traversal of the AMR graph, to process each node to create question-answer pairs.

\subsection{Question Generation}
\label{sec-question-gen}

\begin{figure*}
\centering
\includegraphics[width=0.68\textwidth]{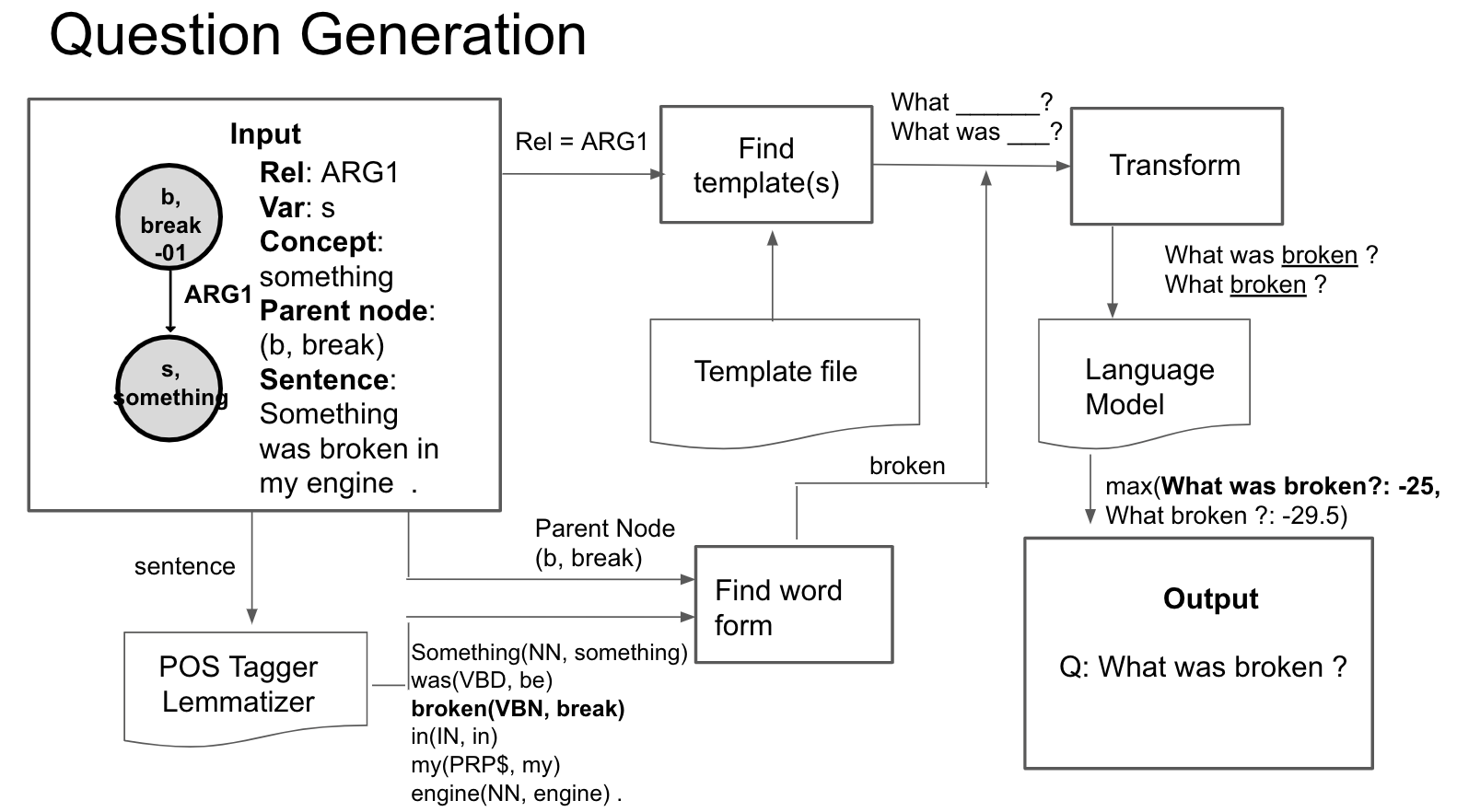}
\caption{Question Generation}
\label{fig-question-gen}
\end{figure*}

We create a question for every node in the final AMR graph, except for the root node. Figure~\ref{fig-question-gen} shows the process of obtaining a question from each node in the AMR graph. 

For each node, we check its relation with its parent, i.e., the label on its link to the parent node, and retrieve the templates for that relation from the previously created templates. For core roles, we also use the parent verb information to select the right templates. The next step is to transform the template(s) into suitable questions, which involves substituting the blank(s) with the correct form of the \textit{concept} from the parent node. This gives us a question pertaining to each template. We score these questions using the GPT-2 language model~\cite{radford2019language}, and output the question with the highest score.

We illustrate the question generation process with an example in figure~\ref{fig-question-gen}. From the input (the sentence, and the node and its parent from its AMR graph), we extract the relation \textit{ARG1}. Next, we extract the templates for \textit{ARG1} : \textit{What was \_\_\_\_\_ ?} and,
\textit{What \_\_\_\_\_  ?} Then, we transform these templates to pose a question for the relation \textit{ARG1}. This requires choosing the right word, i.e., the correct form of the parent concept (in this case break-01) that goes into the blank. We lemmatize the sentence, and get the lemma of each word, to determine which lemma corresponds to the parent concept. "broken" is lemmatized to 'break'. Therefore, we substitute the word "broken" from the sentence into the blank. The two templates are thus transformed to two questions: \textit{What was broken ?}, and \textit{What broken ?}. Of these, \textit{What was broken ?} is chosen as the output because the language model gives it a higher score.
For each verb in the sentence, we also add a question to capture the sense of the verb, which can be answered by the sense annotation.

\subsection{Answer Generation}
\label{sec-answer-gen}

\begin{figure*}
\centering
\includegraphics[width=0.8\textwidth]{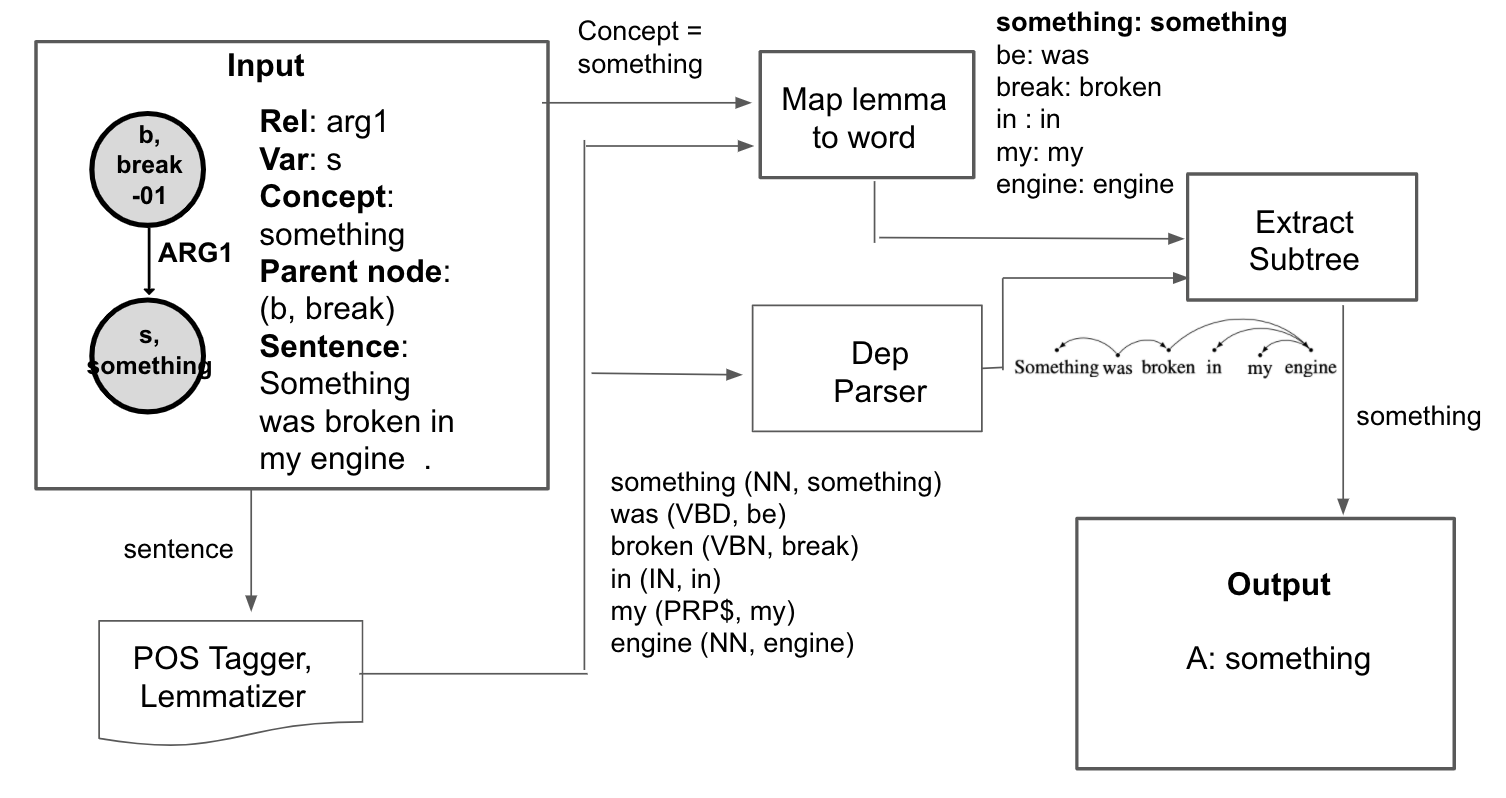}
\caption{The answer generation pipeline.}
\label{fig-answer-gen}
\end{figure*}

The answers to the questions are spans of words from the sentence. The answer could be a single word, or a phrase, which comes from the \textit{concept} of the node for which we are trying to generate a question-answer pair. We lemmatize the sentence to obtain the mapping of concepts in the AMR to words in the sentence. Some questions can be answered by the word to which the \textit{concept} lemma is mapped. However, to extract the correct phrase is not as straightforward. Therefore, we use the Stanza library from~\citet{qi2020stanza} to get the dependency parse tree of the sentence and then extract the sub-tree rooted at the word to which the \textit{concept} lemma is mapped as the answer. Figure~\ref{fig-answer-gen} shows the answer generation module.

\begin{figure}
\centering
\includegraphics[width=0.4\textwidth]{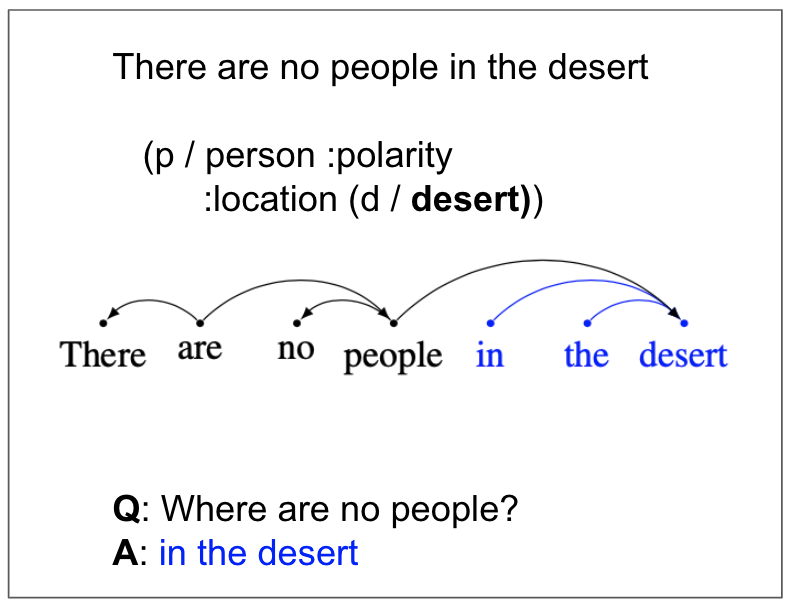}
\caption{Answer extraction from dependency tree}
\label{fig-extract-answer}
\end{figure}

In figure~\ref{fig-extract-answer}, we show an example where we extract an entire phrase to answer a question. This explains how to obtain the answer for a question posed for the node \textit{(d, desert)}. The \textit{concept} \textit{desert} maps to the word \textit{desert}. Therefore, in the dependency parse tree, we look for the head \textit{desert}, and extract all the words modified by \textit{desert} and its children, i.e., all the words from the sub-tree, as the answer.

\section{Experiments and Results}
\label{sec-expts-and-results}

We run ASQ on the AMR annotated Little Prince Corpus (1,562 sentences) and the AMR Annotation Release 2.0 Corpus (39,260 sentences)~\cite{knight2014abstract} to get question-answer pairs.. Table~\ref{tab-stats-lp} show some statistics on the output obtained from both of these datasets. Since they contain gold-standard annotations of AMRs, we perform a qualitative evaluation of the results from AMR 2.0.

\begin{table}
\setlength{\tabcolsep}{2pt} 
\centering
\begin{tabular}{lcc}
\hline
\textbf & \textbf{Little Prince} & \textbf{AMR 2.0} \\ \hline
Total \# of questions & 23,433 & 468,263\\
Avg \# of questions & 15.00 & 11.92 \\
per sentence & & \\
\# of unique words & 12,496 & 183,334 \\
Avg length of questions & 4.23 & 3.92\\ 
Avg length of answers & 3.7 & 2.2 \\\hline
\end{tabular}
\caption{\label{tab-stats-lp}
Statistics for the Little Prince Corpus, and the AMR 2.0 corpus
}
\end{table}

\begin{figure}
\centering
\includegraphics[width=0.50\textwidth]{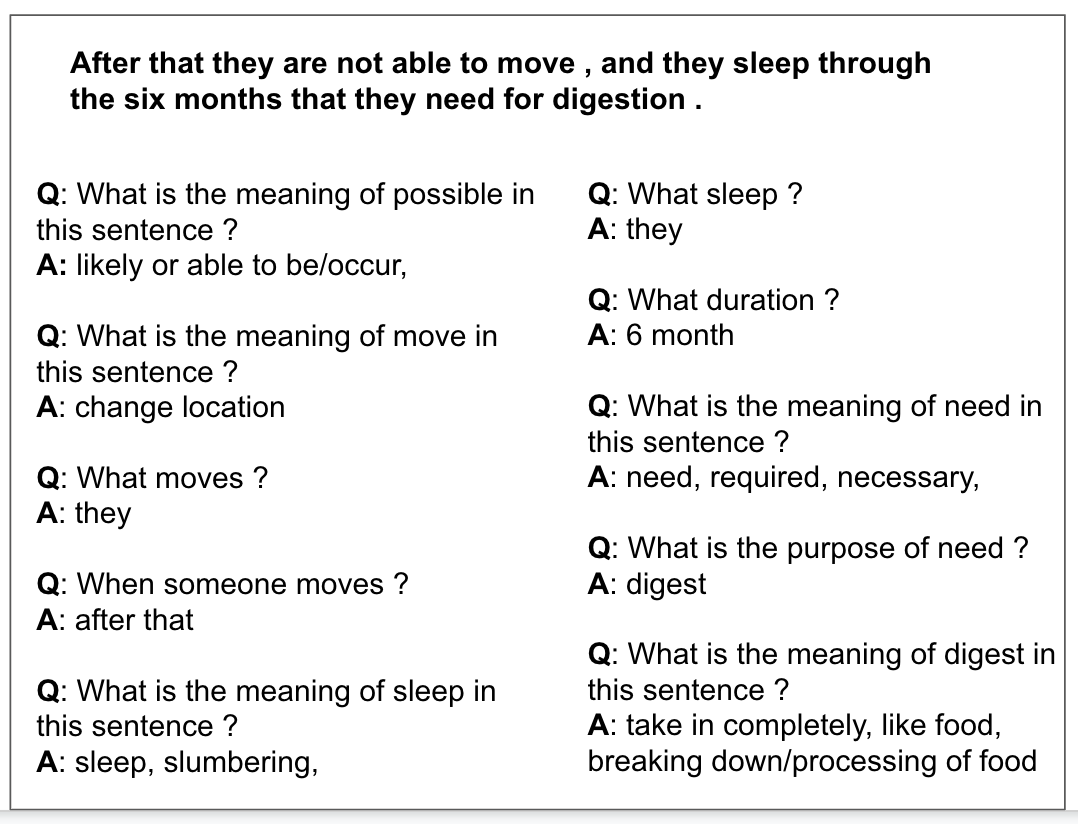}
\caption{Text from the Little Prince Corpus, with output QA pairs from ASQ}
\label{fig-lpr-output}
\end{figure}

\subsection{Human Evaluation}
\label{sec-error-analysis}

We randomly select 70 sentences from the test split of AMR 2.0 and use this as our test set for error analysis.

\paragraph{Qualitative Evaluation}

We show the sentences from the test set and the question-answer pairs generated from them by ASQ to a test subject and ask them to score the quality of each of the QA pairs for grammaticality and naturalness on a scale of 1-5, (5 being the highest), and the correctness of the answer for the question (0 for incorrect, 0.5 somewhat correct, 1 for correct). Also for each sentence, we ask them to rate coverage, i.e., how much information content from the sentence is captured in the set of question-answer pairs generated for it, expressed as a percentage. For the outputs from ASQ, the number of questions is also a good estimate of coverage, since almost all the QA pairs are meant to convey different pieces of information.
These results are summarized in table~\ref{tab-qual-eval}. 

\begin{figure} [ht]
\centering 
\subfloat[\centering QA Pairs from ASQ constructed from gold-labeled AMRs]{{\includegraphics[width=0.50\textwidth]{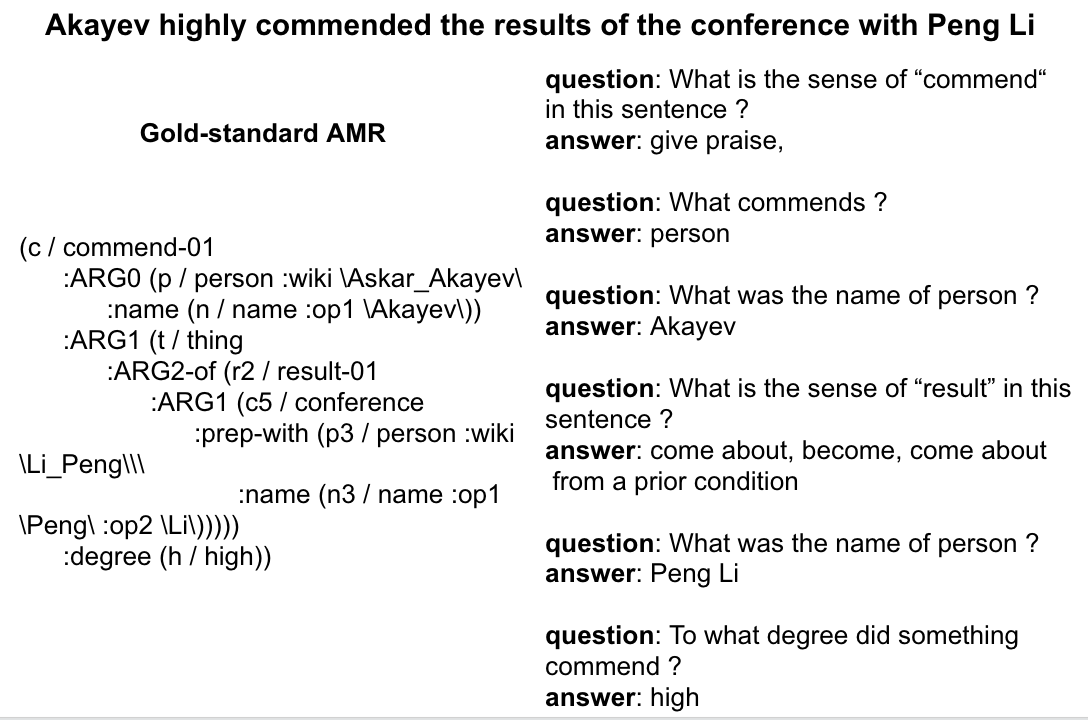} }}%
\qquad 
\subfloat[\centering QA Pairs from ASQ constructed from AMRs predicted by AMR-GS parser]{{\includegraphics[width=0.50\textwidth]{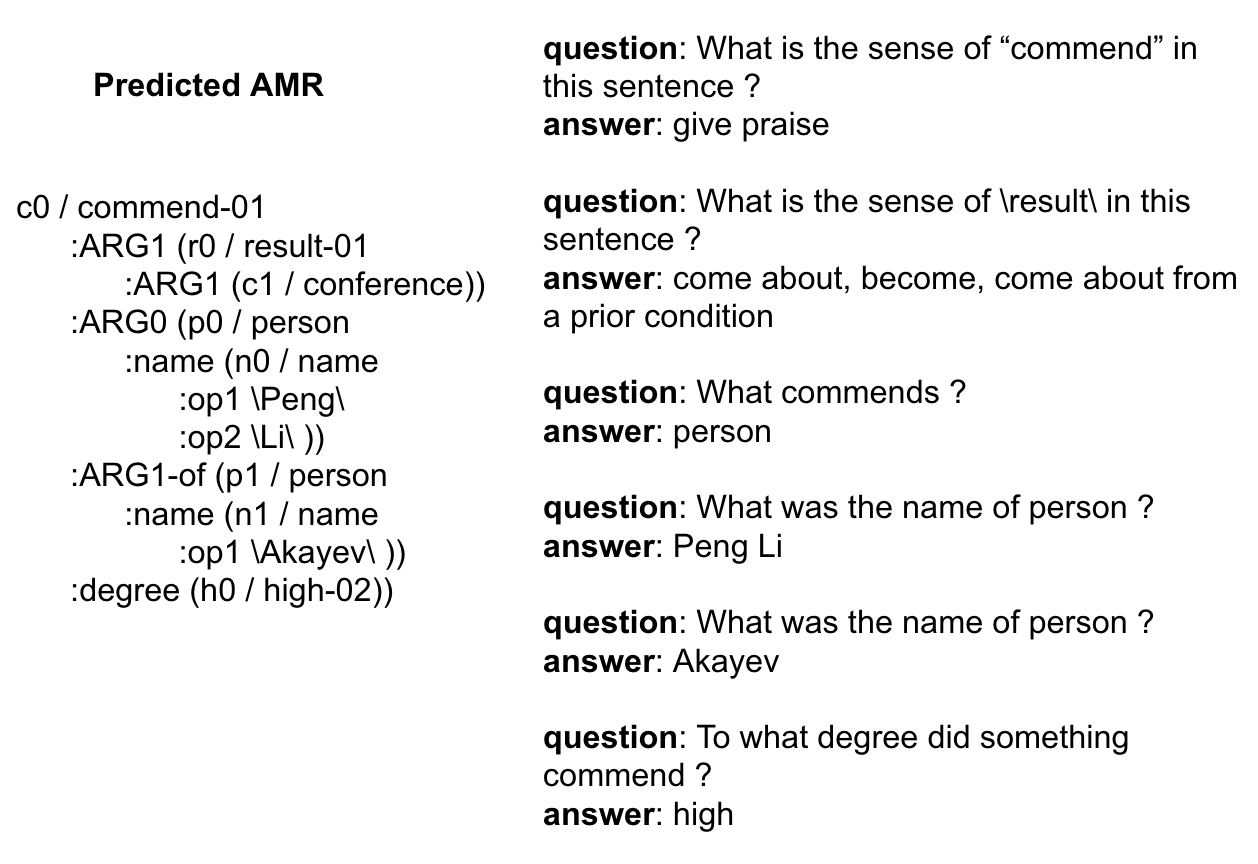} }}%
\caption{For relatively smaller sentences, the output QA pair are similar}%
\label{fig:amr-parsing}%
\end{figure}

\paragraph{Error Propagation from AMR parsers}

We use the AMR-GS parser~\cite{cai2020amr} trained on AMR 2.0, which has close to 80\% accuracy on the AMR2.0 test set. The sentences in QAMR tend to be longer, and more informative, and are from a different domain. The accuracy of the AMR predictions starts to drop as the sentence length increases. For sentences shorter than $12$ tokens, we obtain very similar results for gold standard and predicted AMRs. Figure~\ref{fig:amr-parsing} shows an example of QA pairs from gold standard and predicted AMRs.

\begin{table}
\centering
\begin{tabular}{lrr}
\hline 
Coverage &  69\% \\ 
Grammaticality & 4.1   \\
Naturalness &  3.7    \\  \hline
Percentage of &   67.8\%  \\
correct answers & \\ \hline
\end{tabular}
\caption{\label{tab-qual-eval}
Qualitative Evaluation on a test set of 70 random sentences from the test split of AMR 2.0 showing averaged scores.}
\end{table}

\subsection{Comparison with QAMR}

The QAMR dataset~\cite{michael2017crowdsourcing} contains 100,000 questions from over 5000 sentences, from Wikipedia, WikiNews and the Penn Treebank. These questions were written by annotators manually. In the annotation set up, workers were shown an English sentence with target words highlighted, and they were asked to write \textit{valid} question-answer pairs for each target word. Workers were incentivized to write more questions, to get better coverage. 

We use the same definition of a \textit{valid} question as 
~\citet{michael2017crowdsourcing}, which says a question is \textit{valid} if: it contains at least one word from the sentence; is about the sentence’s meaning; can be answered obviously and explicitly in the sentence; not be a yes/no question; and not be redundant. Redundancy is defined informally as two questions having the same meaning and the same answer. 

Using AMRs to pose questions automatically ensures that the requirements for a question to be \textit{valid} are met. The questions generated are about the meaning of the sentence, and are answerable from information from the sentence. In using AMRs, we automatically ensure maximum coverage, because a good AMR captures most of the factual information in the text. The problem of mapping question-answer pairs to predicate-argument roles becomes redundant, since we create a question-answer pair for almost every relation. Also, there is no scope for redundancy in creating questions from each node of the AMR.

We obtain AMRs for the sentences in the QAMR corpus using the AMR-GS parser~\cite{cai2020amr}. We run ASQ on the sentences from QAMR and their AMRs, to automatically get question-answer pairs, referred to as ASQ-MR. Table~\ref{tab-compare-qamr-asqmr} lists some statistics for the manually created QAMR and automatic ASQ-MR question-answer pairs. The QAMR annotations have higher number of questions per sentence, however, even with filtering, a lot of questions are redundant. Note that the quality of the QA pairs obtained on the QAMR sentences is observed to be much worse than those obtained on the AMR 2.0 or the Little Prince datasets (discussed further in \S\ref{sec-error-analysis}).

\begin{table}
\setlength{\tabcolsep}{2pt} 
\centering
\begin{tabular}{lll}
\hline
\textbf & \textbf{QAMR} & \textbf{ASQ-MR} \\ \hline
Total \# of questions &  89,091 & 56,355 \\
Avg \# of questions &  18.1 & 12.15 \\
per sentence &  &  \\
\# of unique words & 721,122 & 225,420\\
Avg length of questions & 16 & 7.04 \\
Avg length of answers & 6.7 & 3.11 \\  \hline
\end{tabular}
\caption{\label{tab-compare-qamr-asqmr}
Statistics comparing QAMR and automatic ASQ-MR on the QAMR dataset
}
\end{table}

Table~\ref{tab-semantic-roles} shows examples of semantic roles from QAMR and ASQ-MR. One semantic phenomena that the QAMR data exhibits but ASQ-MR doesn't is the use of synonyms to substitute words from the text in creating the questions. It is natural for humans to express information in words that they are more familiar with. This results in more diverse questions in QAMR and a richer vocabulary. 

\begin{table*}[ht]\centering\begin{tabular}{ll}\hline    
\textbf{Semantic role-label} & \textbf{Examples} \\ \hline     
core relations for & His family's high rank enabled Ibn Khaldun to study with the best teachers \\
open-class concepts & in Maghreb.   \\
& \textbf{question}: Who ranks ? \textbf{answer}: His family 's \\ \hline     
modifiers & ASEAN members, together with the group's six major trading partners  \\ 
 &   (Australia, China, India, Japan, New Zealand, South Korea), began the first round \\
 &  of negotiations on 26-28 February 2013, in Bali, Indonesia   on the establishment \\
 &  of the Regional Comprehensive Economic Partnership. \\
& \textbf{question}: What type of partner ? \textbf{answer}: trade \\ \hline    
temporal & Farage, who has lead the party since 2010, announced his resignation on Friday, \\  & with the recommendation that Suzanne Evans serve as interim leader. \\
& \textbf{question}: When someone announces ?  \textbf{answer}: weekday Friday \\ \hline   
location & A petroleum company based in India, Oil and Natural Gas Corporation (ONGC), \\
 & operates in a number of oil blocks under South China Sea with  consent of  \\
  &  Vietnam, in the same Phu Khanh basin. \\
 & \textbf{question}: Where is someone operating ? \textbf{answer}: in the same Phu Khanh basin \\ \hline    
 causality & In the mean time, much stronger friction over land quickly reduces  their strength. \\
 & \textbf{question} Why something reduced ? \textbf{answer}: stronger friction land \\ \hline     
 negation & His friends and family gained important positions without qualifications.\\ 
 & \textbf{question}: How did someone gain?  \textbf{answer}: not qualify \\ \hline     
 quantifiers & People with schizophrenia will sometimes report that, although they are acting  \\
  &  in the  world, they did not initiate, or will, the particular actions they performed. \\
 & \textbf{question}: How often report ? \textbf{answer}: sometimes \\ \hline     
 degree &  The Serbian inventor and electrical engineer Nikola Tesla was heavily influenced  \\ 
 &  by Goethe's Faust, his favorite poem, and had actually memorized the entire text. \\
 & \textbf{question}: To what degree did someone influence ? \textbf{answer}: heavy \\ \hline     
 discourse information &  KH: John and Russ (Nash in the film) went around [scouting] but to be  honest, \\
  & Russ's dad owned the land where we  filmed 90\% of the movie. \\
 & \textbf{question}: What contrast?  \textbf{answer}: KH John went around scouting owned . \\ \hline    
  \end{tabular}\caption{\label{tab-semantic-roles}Examples of semantic roles in ASQ-MR that are also expressed in QAMR}
  \end{table*}

\section{Conclusion and Future Work}
\label{sec:conclusion}

We have presented ASQ, a system for automatically mining questions and their answers from the AMR of a sentence.
This enables construction of QMR datasets automatically in various domains using existing high quality AMR parsers, and provides an automatic mapping AMR to QM for ease of understanding by non-experts.

For future work, it would be possible to improve ASQ by the addition of more templates. ASQ could also be extended to include the use of synonyms. For the data from ASQ, post-editing the questions make an improved version of the dataset. This would require little effort, as it would be easier and quicker for annotators to fix a somewhat incorrect question, than to have to come up with questions by themselves. 

Another possible future direction is to do this generation in the reverse direction, i.e., given a set of question-answer pairs for a sentence, construct its AMR, which is a much more challenging task. The flexibility to switch different meaning representations is yet another step towards understanding language.

\bibliography{emnlp2021}
\bibliographystyle{acl_natbib}

\appendix

\end{document}